# Ranking and Classification-driven Feature Learning for Person Re-identification

Zhiguang Zhang
Beijing University of Posts and Telecommunications
xhx1247786632@bupt.edu.cn

*Abstract*—**Person re-identification has attracted many researchers' attention for its wide application, but it is still a very challenging task because only part of the image information can be used for personnel matching. Most of current methods uses CNN to learn to embeddings that can capture semantic similarity information among data points. Many of the state-of-the-arts methods use complex network structures with multiple branches that fuse multiple features while training or testing, using classification loss, Triplet loss or a combination of the two as loss function. However, the method that using Triplet loss as loss function converges slowly, and the method in which pull features of the same class as close as possible in features space leads to poor feature stability. This paper will combine the ranking motivated structured loss, proposed a new metric learning loss function that make the features of the same class are sparsely distributed into the range of small hyperspheres and the features of different classes are uniformly distributed at a clearly angle. And adopted a new single-branch network structure that only using global feature can also get great performance. The validity of our method is verified on the Market1501 and DukeMTMC-ReID person re-identification datasets. Finally acquires 90.9% rank-1 accuracy and 80.8% mAP on DukeMTMC-reID, 95.3% rank-1 accuracy and 88.7% mAP on Market1501. Codes and models are available in Github. https://github.com/ Qidian213/Ranked_Person_ReID.**

*Keywoeds*—**CNN, Deep metric learning, Image retrieval, Person re-identification.**

## I. INTRODUCTION

With the development of deep learning, person re-identification based on deep metric learning has also achieved good performance. However, many state-of-the-arts methods use complex network structures with multiple branches that fuse multiple features[6][7][10], that is very inconvenient to use. To learn embeddings that can capture semantic similarity information among data points, a lot of loss function have been proposed int the literature, such as Contrastive loss[18],Triplet loss[21]. Fo r the outperform performance of Triplet loss, it was widely used. But using Triplet loss as loss function the method in which pull features of the same class as close as possible in features space leads to poor feature stability. Although ranking motivated structured loss[1] has relieve the problem of feature stability, but the use of difficult samples is still insufficient, and only focus on the feature distribution of the same class ignore the optimization of the feature distribution between different classes.

This paper combine the ranking motivated structured losses and classification loss, not only pay attention to the distribution of features within class to ensure the stability but also the distribution of features between classes. So that the distribution of features between different classes are more distinguishable. In addition, a single-branch structure is used to ensure the convenience of use. Experiments show that our method achieved good performance.

The key contributions of this paper are as follows:
- Do more research on ranking motivated structured loss and combined Softmax loss(LS) design a new loss function.
- Adopted a simple single-branch network structure achieved good performance.

The paper is organized as follows. In Section 2, we present the background and motivation. In Section 3, we describe the loss function and netwrok we proposed. In Section 4, firstly we give the test datasets and methods, and then analyze the experiment results. Finally, In Section 5, we discuss our method and then draw some conclusions.

## II. REALATED WORK

Loss functions currently used for metric learning mainly include classification loss such as Softmax loss and metric learning loss such as Contrastive loss [18] and Triplet Loss[21]. Softmax loss is suitable for traditional classification, such as the classification and identification of flowers and birds, which can make the features of different class clearly separated by a certain angle, but there is a case where the distance between classes is smaller than the distance within the class, as shown in Fig. 1(a). And to prevent overfitting for a classification task while training, Label smoothing (LS) proposed in [22] and can significantly improve the performance of the model.

Like Triplet Loss[19] metric learning losses are suitable for fine-grained identification at the individual level, such as identifying a person's identification in a group of people. Given a query the initial Triplet loss and the improved version Hard Triplet loss [17] only use the nearest negative sample and the farthest positive sample, which is not robust to outliers and cannot make full use of the hard samples. Later, Adaptive Weighted Triplet Loss [20] adding positive weights to the positive and negative samples based on their distance from the query, but still not solve the fundamental disadvantage of Triplet loss, the method in which pull features of the same class as close as possible in features space leads to poor feature stability and convergence slowly, as shown in Fig.1(b). In [1], the author proposed Ranked list loss based on the ranking motivated structured losses, proposed to learn a hypersphere for each class in order to preserve the similarity structure inside it, as shown in Fig.1(c). For a query sample, when the positive samples's features are distributed into a hypersphere with a certain radius, the corresponding positive samples are no longer optimized, when the negative samples's features are distributed outside of a bigger hypersphere with a certain radius, the corresponding negative samples are no longer optimized. And, according to the distance of the negative sample from the big hypersphere surface, the corresponding weight is calculated by using exponential function to utilize more negative samples's information. This avoids the destruction of features. However, Ranked list loss mainly pays attention to the distribution of features intraclass to avoid the destruction of features, does not pay attention to the distribution of features between classes. The feature distribution between classes is unconstrained and almost random. And the radius of big hypersphere set for negative samples is small, so the weight difference between the easy negative samples and the difficult negative samples is small leads the utilization of the difficult negative sample insufficient.

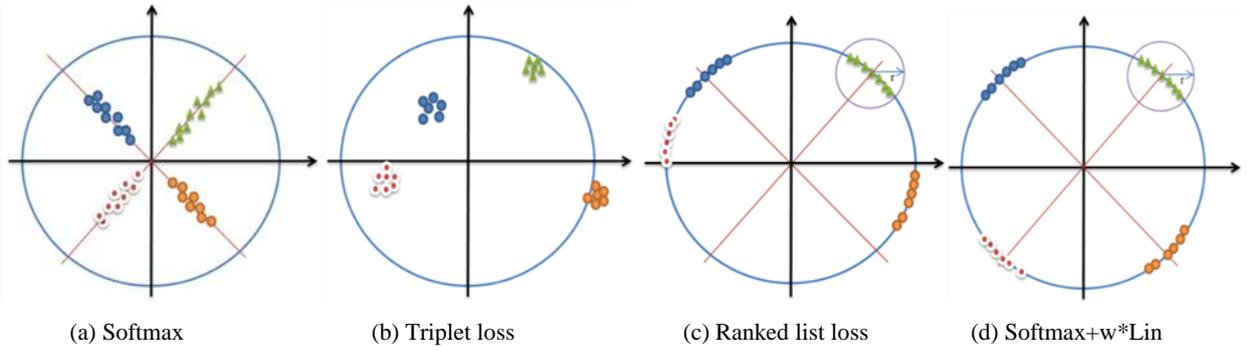

(a) Softmax  (b) Triplet loss  (c) Ranked list loss  (d) Softmax+w*Lin

Fig.1. Illustration of different losses. Different color represent different classes. (a)(b)(c)(d) represent the feature distribution when use corresponding loss function.

For the network, many state-of-the-arts methods use complex network structures with multiple branches that fuse multiple features while training or testing, this is very inconvenient to use. DCDS[25] use two branches combined different level features, MMGA[26] use a multi-scale body-part mask guided attention network, the training of attention modules is guided by whole-body masks, upper-body masks and bottom-body masks. OSNET[27] with the layer depth incremented across different streams to achieve different scales, dynamically aggregated to generate omni-scale features. A relatively simple network structure were proposed in Bnnk[13], and achieves state-of-the-arts performance.

In order to overcome the problems of Ranked list loss and pay attention to the distribution of features between classes. we designed a new loss function to ensure the features of person image more distinguishable. And in order to meet the purpose of convenient use, we adopted a single-branch network structure.

### III. METHOD

In this section, we describe the method we proposed. Firstly introduce the loss function we designed then is a single-branch network structure.

### 3.1 Loss function

Inspired by Ranked list loss , our objective is to learn a discriminative function f (a.k.a. deep metric) that make the features of the same class sparsely distributed into the range of small hyperspheres and the features of different classes are uniformly distributed at a clearly angle, as shown in Fig.1(d). The design of Lin loss function are based on Ranked list loss, and do some change for our purpose.

*1). Pairwise Constraint*

For the image xi to be queried, the objective is optimize the positive sample into a hyperspheres which center is the feature of xi radius is r, the negative sample is optimize to have the maximum distance on a big hypersphere surface. When the features are normalized, the maximum distance between two features is 2, so the maximum distance in this paper is 2. The loss function designed for image pair is:

$$Lm(x_i, x_j; f) = (1 - y_{ij})[2 - d_{ij}]_+ + y_{ij}[d_{ij} - r]_+ \quad (1)$$

where $y_{ij} = 1$ if $y_i = y_j$, otherwise $y_{ij} = 0$, $d_{ij}=\|f(x_i)-f(x_j)\|_2$ is euclidean distance between two features.

*2). Loss for positive samples*

For positive samples to xi, when the distance greater than r will be regarded as difficult positive samples. The loss function is Eq.(2), Np is number of positive samples.

$$Lp(xi; f) = \frac{1}{N_P} \sum Lm(x_i, x_j; f) \quad (2)$$

*3). Loss for negative samples*

For negative samples to xi, due to the relatively large number of positive samples. To make better use them, especially difficult negative samples, give every negative sample a weight based on the difference between their distance to xi and the maximum distance between two features:

$$W_{ij} = exp(-d\_ij) \cdot exp(T \cdot (2 - d_{ij})) \quad (3)$$

Different from Ranked list loss in the loss function, value 2 is the max distance between two features and add scale by negative parameter exponential function, this let every negative sample to participate in the loss calculation. According to the characteristics of the exponential function, when the difference from difficult negative sample to target and simple negative sample to target is determined, if the difference to the optimization target become bigger, the ratio of the weight for difficult negative sample and simple negative sample will becomes larger. This can use more difficult negative samples's information, less use simple negative samples's information to avoid damage to features of simple negative samples. Where T controls the degree of weighting negative examples and the gradient coefficient of negative sample. Finally, all negative samples in the input data are combined in a weight normalization manner:

$$Ln(x_i; f) = \sum \frac{w_{ij}}{\sum w_{ij}} Lm(x_i, x_j; f) \quad (4)$$

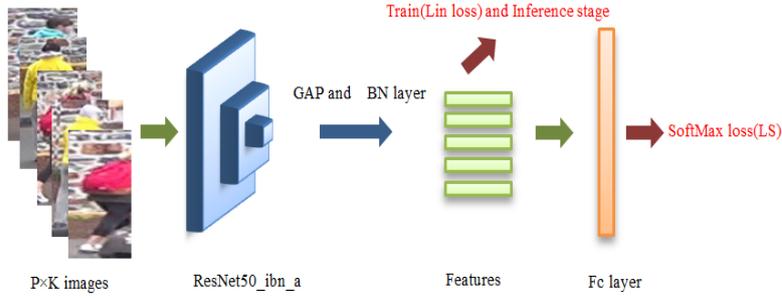

Fig. 2. The pipeline used in this paper

Finally, the loss function used to optimize the features intraclass is :

$$Lin = Lp + Ln \quad (5)$$

*4). Overall of loss function*

The above loss function focuses on the distribution of features intraclass, the distribution of features between classes is still not well optimized. In order to ensure that the features of different classes are better differentiated. Combine Softmax loss and Lin loss as the final loss function , w is the weight of Lin loss function. To prevent overfitting we use Softmax loss with Label smoothing.

$$M\_loss = Softmax(LS) + w \cdot Lin \quad (6)$$

### 3.2 Network

In order to meet the purpose of convenient use, proposed a single-branch network structure, that add BN layer after average pooling layer of backbone network then is a fully connected layer, and the bias of the BN layer and FC layer are removed. Lin loss training and inference stage all use features after the BN layer, Softmax(LS) is used after FC layer, and the features vector after BN is normalized modulus is 1. For backbone network we use Resnet50_ibn_a proposed by[23], which has same parameters and computational cost with Resnet50, performance well on semantic segmentation task. The author find with the depth of the network increases, the differences in the features of different dataset become smaller and smaller, the differences in the characteristics of different categories become larger and larger and concludes the role of IN and BN layers in CNNs: IN introduces appearance invariance and improves generalization while BN preserves content information in discriminative features, and proposed IBN-Net, which carefully unifies instance normalization and batch normalization layers in a single deep network to increase both modeling and generalization capacity without parameters and computational cost increase. The network as shown in Fig. 2.

## IV. EXPERIMENT

In this section, we conduct experiments on two popular benchmarks: Market1501[14]，DukeMTMC-ReID[16]. And use training tricks adopted by Bnnk, like set last stride to 1, warmup learning rate, random erasing augmentation, do not change any training settings. For mini-batchsizes we randomly sample P identities and K images of per person. Finally the batch size equals to B = P×K. In this paper, we set P = 16 and K = 4.

### 4.1 Baseline

For comparison , experiments were conducted use Lin loss or Softmax Loss(LS) alone, and Lin loss was experiment separately to train using features before and after BN layer, Softmax was experiment separately to train with BN layer or not.

**TBALE I**
RESULT OF USE DIFFERENT FEATURES FOR TRAINING, EG: LIN_BEFORE MEANS USE FEATURES BEFORE BN LAYER, SOFTMAX_BN MEANS ADD BN LAYER TO NETWORK.

| Loss | r | T | DukeMTMC-ReID | | Market1501 | |
|---|---|---|---|---|---|---|
| | | | mAP | rank1 | mAP | rank1 |
| Lin_before | 0.7 | 1.0 | 29.5 | 46.2 | 40.1 | 58.3 |
| Lin_after | 0.7 | 1.0 | **57.4** | **73.6** | **65.3** | **80.1** |
| Softmax(LS) | | | 77.3 | 88.2 | 85.6 | 94.1 |
| Softmax_BN(LS) | | | **77.3** | **88.5** | **86.1** | **94.6** |

### 4.2 Experiments of our method

**TABEL II**
PERFORMANCE OF DIFFERENT T VALUES.

| w | r | T | DukeMTMC-ReID | | Market1501 | |
|---|---|---|---|---|---|---|
| | | | mAP | rank1 | mAP | rank1 |
| 0.4 | 0.7 | 0.5 | 80.4 | 90.6 | 88.7 | **95.1** |
| 0.4 | 0.7 | 1.0 | **80.8** | **90.9** | **88.9** | 94.9 |
| 0.4 | 0.7 | 5.0 | 80.3 | 90.6 | 88.5 | 94.7 |

**TABEL III**
PERFORMANCE OF DIFFERENT R VALUES.

| w | r | T | DukeMTMC-ReID | | Market1501 | |
|---|---|---|---|---|---|---|
| | | | mAP | rank1 | mAP | rank1 |
| 0.4 | 0.6 | 1.0 | 80.3 | 90.6 | 88.7 | 94.6 |
| 0.4 | 0.7 | 1.0 | **80.8** | **90.9** | **88.9** | **94.9** |
| 0.4 | 0.8 | 1.0 | 80.8 | 90.4 | 88.8 | 94.8 |

**TABEL IV**
PERFORMANCE OF DIFFERENT W VALUES

| w | r | T | DukeMTMC-ReID | | Market1501 | |
|---|---|---|---|---|---|---|
| | | | mAP | rank1 | mAP | rank1 |
| 0.2 | 0.7 | 1.0 | 80.2 | 90.1 | 88.7 | **95.3** |
| 0.4 | 0.7 | 1.0 | **80.8** | **90.9** | 88.9 | 94.9 |
| 0.6 | 0.7 | 1.0 | 80.6 | 90.2 | **89.0** | 95.1 |

Using Eq.(6) as loss function. Using features after BN layer to compute Softmax loss(LS) and Lin loss, test the performance using features after BN layer. And experimental analysis the influence of different w, r, t values. Shown as Table II-IV.

The parameter w controls the weight of optimizate the feature distribution intraclass and the feature distribution between classes. r controls the optimization space of the features within the class and play a significant role in the stability of feature. T determines the weight of the negative sample and the scaling of gradient. Parameters w, r affects the performance of the model more than T.

### 4.3 Experiments on different backbone

To further explore effectiveness, we also present the results of use different backbone network in Table V.

TABLE V
PERFORMANCE OF DIFFERENT BACKBONE NETWORK

| Network | DukeMTMC-ReID | | Market1501 | |
|---|---|---|---|---|
| | mAP | rank1 | mAP | rank1 |
| ResNet34 | 74.9 | 87.0 | 84.5 | 93.7 |
| ResNet50 | 77.9 | 88.5 | 87.2 | 95.0 |
| Se_ResNet50 | 78.0 | 88.3 | 87.7 | 94.5 |
| Se_ResNext50 | 79.0 | 89.5 | 88.6 | 95.3 |
| ResNet50_ibn_a | 80.8 | 90.9 | 88.7 | 95.3 |

### 4.4 Cross domain

To further explore effectiveness, we also present the results of cross-domain experiments in Table VI.

TABEL VI
THE PERFORMANCE OF DIFFERENT MODELS IS EVALUATED ON CROSS-DOMAIN DATASETS. M->D MEANS THAT WE TRAIN THE MODEL ON MARKET1501 AND EVALUATE IT ON DUKEMTMC-REID.

| D->M | | M->D | |
|---|---|---|---|
| mAP | rank1 | mAP | rank1 |
| 25.5 | 52.7 | 28.6 | 45.5 |

### 4.5 Comparison of State-of-the-Arts

We compared our method with the following methods which are with State-of-the-Arts performance or new proposed.

TABEL VII
COMPARISON OF STATE-OF-THE-ARTS METHODS. NF IS THE NUMBER OF FEATURES WHEN TEST. RK STANDS FOR K-RECIPROCAL RE-RANKING METHOD [24]

| Method | Nf | DukeMTMC-ReID | | Market1501 | |
|---|---|---|---|---|---|
| | | mAP | rank1 | mAP | rank1 |
| SPReID[2] | 5 | 71.0 | 84.4 | 81.3 | 92.5 |
| MaskReID[3] | 3 | 61.9 | 78.8 | 75.3 | 90.0 |
| SCPNet[5] | 1 | 62.6 | 80.3 | 75.2 | 91.2 |
| PCB[6] | 6 | 69.2 | 83.3 | 81.6 | 93.8 |
| Pyramid[7] | 1 | - | - | 82.1 | 92.8 |
| Mancs[8] | 1 | 71.8 | 84.9 | 82.3 | 93.1 |
| DuATM[9] | 1 | 62.3 | 81.2 | 76.6 | 91.4 |
| HA-CNN[10] | 4 | 63.8 | 80.5 | 75.7 | 91.2 |
| SVDNet[11] | 1 | 56.8 | 76.7 | 62.1 | 82.3 |
| AWTL[12] | 1 | 63.4 | 79.8 | 75.7 | 89.5 |
| BNNK[13] | 1 | 76.4 | 86.4 | 85.7 | 94.1 |
| DCDS[25] | 2 | 75.5 | 86.1 | 85.8 | 94.1 |
| MMGA[26] | 1 | 78.1 | 89.5 | 87.2 | 95.0 |
| OSNET[27] | 1 | 73.5 | 88.6 | 84.9 | 94.8 |
| Ours | 1 | **80.8** | **90.9** | **88.7** | **95.3** |
| Ours(RK) | 1 | **90.0** | **92.6** | **94.4** | **95.7** |

### 4.6 Analysis

In experiment 4.1, use Lin loss or Softmax loss(LS) alone can only get general performance on rank1 or mAP, and use Softmax loss(LS) performance better. Think that Lin loss mianly constrains the distribution of features intraclass, it cannot guarantee that the distribution of features between classes is better optimized. Combined with experiment 4.2, features after BN layer can get better results and Lin loss can help get better results. Think that BN layer can convert the values of feature vector of samples into normal distribution, avoid large values in feature vector, outlier coordinate representation, avoid Fc layer to have large weights in order to fit a certain class's features, this can prevent overfitting and BN layer makes the cost space smoother can helps optimizate model. And BN layer associating different samples increases globally optimization. And combined Lin loss and Softmax loss got better performance, think that Softmax loss(LS) provide globally optimal constraint makes features of different class distributed at a clearly angle, Lin loss provide intraclass optimal constrain makes the features of each class sparsely distributed in different subspaces and maintain the stability of the feature. And think to optimizate the model Softmax loss(LS) plays a key role, Lin loss plays a supporting role, as shown in table IV, w can only set small value.

## V. CONCLUSION

In this paper, we do more research on Ranked list loss and combined Softmax loss(LS) design a new loss function, and to ensure the convenience of use we adopted a single-branch network structure that only using global feature for training and testing, and a lot of experiments verified its effectiveness. Finally we achieved outstanding performance.